\pdfoutput=1

\documentclass[11pt]{article}

\usepackage[]{acl}

\usepackage{times}
\usepackage{latexsym}

\usepackage[T1]{fontenc}

\usepackage[utf8]{inputenc}

\usepackage{microtype}

\usepackage{inconsolata}

\usepackage{graphicx}
\usepackage{multirow}
\usepackage{booktabs}
\usepackage{subfig}
\usepackage{enumitem}

\newcommand{\dataName}{\textsc{ReCo}}

\title{Can LLMs Estimate Cognitive Complexity\\ of Reading Comprehension Items?}

\author{
    Seonjeong Hwang$^1$, 
    Hyounghun Kim$^{1,2}$, 
    Gary Geunbae Lee$^{1,2}$ \\
    $^1$Graduate School of Artificial Intelligence, POSTECH, Republic of Korea\\
    $^2$Department of Computer Science and Engineering, POSTECH, Republic of Korea\\
    \texttt{\{seonjeongh,
    h.kim, 
    gblee\}@postech.ac.kr} \\
}

\begin{document}
\maketitle
\begin{abstract}
Estimating the cognitive complexity of reading comprehension (RC) items is crucial for assessing item difficulty before it is administered to learners.
Unlike syntactic and semantic features, such as passage length or semantic similarity between options, cognitive features that arise during answer reasoning are not readily extractable using existing NLP tools and have traditionally relied on human annotation.
In this study, we examine whether large language models (LLMs) can estimate the cognitive complexity of RC items by focusing on two dimensions—Evidence Scope and Transformation Level—that indicate the degree of cognitive burden involved in reasoning about the answer.
Our experimental results demonstrate that LLMs can approximate the cognitive complexity of items, indicating their potential as tools for prior difficulty analysis.
Further analysis reveals a gap between LLMs’ reasoning ability and their metacognitive awareness: even when they produce correct answers, they sometimes fail to correctly identify the features underlying their own reasoning process.\footnote{The dataset, prompt templates, and evaluation codes are
available at \url{https://github.com/SeonjeongHwang/ReCo}.}

\end{abstract}

\section{\label{sec:intro} Introduction}

\begin{figure}[ht]
\centering
\includegraphics[width=\columnwidth]{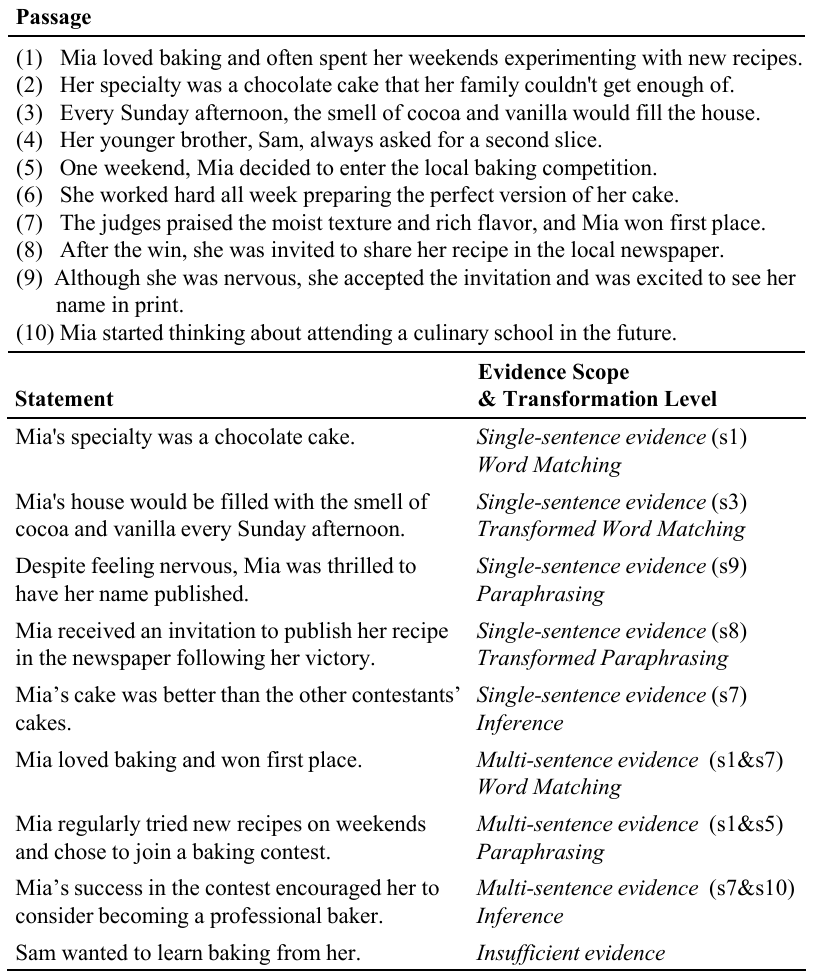}
\caption{\label{fig:main_figure} Examples of RC items that require determining the factuality of a statement. Each item is annotated along two cognitively grounded dimensions (\textit{Evidence Scope} and \textit{Transformation Level}) with corresponding supporting sentences highlighted from the passage.}
\end{figure}

Estimating the difficulty of reading comprehension (RC) items\footnote{An RC item typically consists of a passage, a question stem, and sometimes answer options (in the case of the multiple-choice format). In this paper, we use the term “item” interchangeably with “question.”} is essential for delivering appropriate learning materials and constructing balanced test forms.
Traditionally, difficulty has been derived from student responses using frameworks such as classical test theory (CTT) or item response theory (IRT)~\cite{lord1980irt,hambleton1993comparison}.
However, these approaches are only applicable after test administration and thus cannot support prior difficulty prediction during item development.
Expert judgment remains the common alternative, but it is costly, time-consuming, and subject to rater variability~\cite{alkhuzaey2024text}.
Another line of work has attempted to simulate student responses using deep learning models of varying capability~\cite{lalor2019learning,uto2023difficulty,park2024large}, thereby enabling difficulty estimation without real student data.
However, this approach incurs substantial computational costs for training or running inference over many models, and the resulting difficulty estimates lack interpretability in terms of what makes an item difficult.

In parallel, researchers have explored a complementary direction: leveraging item features to understand the relationship between item characteristics and difficulty in a more interpretable way~\cite{pandarova2019predicting,choi2020predicting,benedetto2021application,byrd2022predicting}.
Syntactic, semantic, and psycholinguistic variables (e.g., sentence length, semantic similarity between options, and word familiarity) have been predominantly leveraged for this analysis, yet these text-based features provide limited insight into the reasoning processes that largely govern difficulty.

Educational psychology research has shown that cognitive factors involved in the answer decision process are more strongly associated with difficulty~\cite{embretson1987component}. 
Examples include the amount of text that must be referenced to determine the correct answer and the degree of transformation between passage evidence and the answer~\cite{bormuth1970children,anderson1972construct}. 
However, these cognitive features cannot be automatically extracted with existing NLP tools, and prior studies have relied exclusively on human raters~\cite{hutzler2014learning,lai2017race}.
This raises a key challenge: \emph{How can we estimate the cognitive complexity of RC items in a scalable way?}
We believe that large language models (LLMs), with their powerful reasoning and instruction-following capabilities, may offer a promising approach to this problem.

Recent studies have attempted to leverage LLMs for estimating question difficulty.
However, many of these efforts have focused on tasks solvable solely through a model’s internal knowledge—such as mathematics and coding~\cite{rogoz2024unibucllm,park2024large,ko2024hierarchical,xu2024adaption}—or have directly prompted LLMs to predict the difficulty of RC items~\cite{raina2024question}.
Whether LLMs can meaningfully analyze the cognitive complexity involved in solving RC items, however, remains largely unexplored.

Motivated by this gap, we investigate LLMs’ capability to measure the complexity of two cognitive variables: \textbf{Evidence Scope} and \textbf{Transformation Level}.
Evidence Scope reflects the amount of text required to verify an answer---categorized as single-sentence, multi-sentence, or insufficient---while Transformation Level captures the degree of lexical and structural transformation between an option and its supporting evidence in the passage, ranging from word matching to inference.
To support empirical evaluation, we constructed \textbf{\dataName}, a benchmark reading comprehension dataset with cognitive complexity annotations, comprising 776 RC items annotated along these two cognitive dimensions (see Figure~\ref{fig:main_figure}). 

In our experiments, we evaluate eight LLMs, spanning both proprietary and open-source variants.
The results show that LLMs can approximate cognitive complexity, with the best-performing models achieving Macro F1 scores of 74.8 (Evidence Scope) and 82.0 (3-level Transformation Level) in the cognitive complexity classification tasks.
Notably, open-source models such as Qwen2.5 (32B) and Mistral-Small (24B) performed comparably to—or even surpassed—GPT-4o.
However, we also found that LLMs often struggle to explicitly recognize key features in their reasoning traces—for example, identifying phrase reordering or the evidence sentences they referenced—highlighting a gap between reasoning ability and metacognitive awareness.

Our contributions can be summarized as follows:
\begin{itemize}[nosep]
    \item We construct an expert-annotated dataset of RC items along two cognitively grounded dimensions, which are important factors for analyzing item difficulty.
    \item We conduct an evaluation of eight instruction-tuned LLMs, demonstrating their potential utility in estimating the cognitive complexity of RC items.
    \item We probe LLMs on fine-grained cognitive features and observe that, even when they successfully solve items, they do not fully recognize the cognitive processes underlying their problem solving.
\end{itemize}

\section{\label{sec:related_work}Related Work}

\subsection{Difficulty Factors and Taxonomies}

Research in educational psychology has long examined the factors influencing RC item difficulty~\cite{bormuth1970children,anderson1972construct,freedle1991prediction,park2004comparison,rafatbakhsh2023predicting}. 
Researchers have sought to identify correlations between item-based attributes—such as surface-level linguistic features (e.g., sentence complexity, vocabulary difficulty) and cognitive burden factors (e.g., plausibility of distractors, option–text mapping)—and item difficulty, often measured using CTT or IRT~\cite{hsu2018automated,pandarova2019predicting,choi2020predicting,zhou2020multi,benedetto2021application}.
More recently, tools such as Coh-Metrix~\cite{mcnamara2014automated}, NLTK~\cite{loper2002nltk}, and embedding models like Word2Vec~\cite{mikolov2013efficient} or BERT~\cite{devlin2019bert} enable the automatic extraction of syntactic and semantic features.
These features have been used as inputs to supervised models such as linear regression for difficulty prediction.

Several taxonomies have been proposed to systematically classify the complexity of RC items.
Bloom’s taxonomy, for instance, organizes learning objectives by levels of cognitive demand~\cite{bloom1956taxonomy}.
\citet{lai2017race} employed a five-level taxonomy to characterize reasoning across RC items, which is a simplified combination of the two dimensions adopted in our study.
While previous taxonomies have been used to categorize different types of RC items—such as main idea, author’s intent, fill-in-the-blank, and detail information questions—our study focuses on distinguishing variations in cognitive complexity, even within items of the same type.

\subsection{LLM-based Difficulty Estimation}

Recently, various approaches have been proposed to leverage LLMs for predicting item difficulty.
Some studies directly prompted LLMs to estimate difficulty~\cite{xu2024adaption}, while others inferred difficulty from model-generated outcomes such as answering accuracy or confidence scores~\cite{rogoz2024unibucllm,park2024large,lu2024generative,jain2025exploring}.
However, much of this prior work has focused on domains such as mathematics, medicine, or coding, where models rely solely on their internal knowledge to solve the problems.
This differs from RC, where the model must reference the information provided in the passage while applying its own reasoning ability.

Several studies have explored LLMs for predicting RC item difficulty.
\citet{raina2024question} found that comparative prompting—asking an LLM to compare the relative difficulty of two items—aligned better with human judgments than absolute prompting, where the model assigns a difficulty score to a single item.
\citet{dutulescu2024hard} predicted item difficulty using indicators derived from LLMs’ question answering (QA) loss.
\citet{kapoor2025prediction} showed that combining item text features, LLM embeddings, and contextual information (e.g., grade level, year) improved prediction performance, underscoring the importance of item-feature-based analyses.
Instead of directly predicting item difficulty, this paper investigates whether LLMs can estimate two cognitively grounded features that influence difficulty.

\subsection{Datasets}

While QA datasets such as SQuAD~\cite{rajpurkar2016squad} and BoolQ~\cite{clark2019boolq} are widely used, they lack per-item difficulty annotations, limiting their usefulness for difficulty analysis.
RACE++~\cite{lai2017race,liang2019new} contains RC items spanning middle school through college levels and has been used with grade level as a proxy for difficulty~\cite{raina2024question,liusie2023analysis}, but this approach does not capture fine-grained variation within a single learner group.
Multi-hop QA datasets such as HotpotQA~\cite{yang2018hotpotqa} assess complexity through multi-hop reasoning across documents, yet this setup differs from standard RC formats in educational assessment, which typically involve a single reading passage.

\citet{huang2017question} predicted item difficulty using student error rates on English reading problems in China, but the full dataset was not released.
\citet{mullooly2023cambridge} released the CMCQRD dataset, which contains 289 RC items labeled with CEFR levels and IRT-based difficulty scores derived from pretesting, providing holistic estimates of item difficulty.
\citet{dutulescu2024hard} annotated FairytaleQA~\cite{xu2022fantastic} along two dimensions: explicit vs. implicit and local vs. summary.
This dataset is the closest to our ours, but the two dimensions considered in this work capture more fine-grained cognitive features involved in the answer decision process of RC items.

\section{\label{sec:data_consruction} Data Construction}

To construct our dataset, we used True/False/Not Given (TFNG) items, where the task is to assess the factuality of a statement given a passage.
Each item comprises a reading passage and a declarative statement, as illustrated in Figure~\ref{fig:main_figure}.
This format is particularly suitable for our study, as it spans a wide range of cognitive complexity—from direct span matching to multi-sentence inference—and is commonly featured in the RC sections of standardized proficiency exams.

\subsection{\label{sec:dimension_descrition} Dimensions of Cognitive Complexity}

\paragraph{Evidence Scope.}
Items that can be solved by referencing a single sentence in the passage are generally easier than those requiring integration of information scattered across multiple sentences~\cite{bormuth1970children,park2004comparison}.
In this study, Evidence Scope refers to the span of text required to determine the truth value of a statement, and is categorized into three levels:
\begin{itemize}[nosep]
    \item \textbf{Single-sentence evidence}: All necessary information to evaluate the statement is contained within a single sentence in the passage.
    \item \textbf{Multi-sentence evidence}: The required information is distributed across multiple sentences (i.e., inter-sentence comprehension).
    \item \textbf{Insufficient evidence}: The passage lacks adequate information to definitively confirm or refute the statement. In such cases, learners are required to examine the entire passage before concluding that the passage provides no supporting evidence.
\end{itemize}
A special case arises when the supporting evidence includes anaphoric expressions.
While \citet{bormuth1970children} treated such items as a separate category, they found little difference in difficulty compared to single-sentence evidence.
Accordingly, we apply the following rules:
1) If the anaphora clearly refers to a frequently mentioned and easily identifiable entity in the prior sentences, the item is classified as single-sentence evidence.
2) However, if resolving the anaphora requires referring back to a prior sentence, we label it as multi-sentence evidence.
This approach reflects that many reading passages in RC assessments employ anaphoric references, and accounting for such subtleties is essential for accurate difficulty prediction.

\paragraph{Transformation Level.}
When the degree of transformation between a statement and its supporting evidence is higher, identifying the corresponding passage text and assessing the statement’s truth value imposes greater cognitive demands~\cite{bormuth1970children,anderson1972construct}.
We adopt a \textit{5-level} taxonomy inspired by previous work, which captures the type of transformation required to derive a statement from the evidence:
\begin{itemize}[nosep]
    \item \textbf{Word Matching}: The content words in the statement appear verbatim in the evidence, and the phrase order is preserved.
    \item \textbf{Transformed Word Matching}: The content words are still present in the evidence but have been rearranged.
    \item \textbf{Paraphrasing}: The statement rephrases the content words without changing the order of the words.
    \item \textbf{Transformed Paraphrasing}: The content words are rephrased and the phrase order is altered, combining lexical and structural transformation.
    \item \textbf{Inference}: The statement cannot be directly derived from any surface form in the passage, even through paraphrasing or reordering; instead, it requires inference.
\end{itemize}
In contrast to the \textit{single-sentence evidence} cases, phrase reordering is either trivial or pervasive in \textit{multi-sentence} cases; therefore, we label these items using a simplified \textit{3-level} taxonomy: \textit{word matching}, \textit{paraphrasing}, and \textit{inference}.
In addition, anaphora resolution, identifying the antecedent of an anaphor within a text, is not considered paraphrasing unless additional lexical transformation is involved.

\subsection{\label{data_construction} Data Annotation}

RACE++~\cite{lai2017race,liang2019new} is a RC dataset consisting of English RC items sourced from exams administered to Chinese middle school, high school, and college students.
To construct \dataName, we use Multiple-choice True/False (MTF) items from the RACE++, each consisting of a reading passage and four options.
We collected only items requiring holistic passage comprehension, excluding those targeting specific entities or local paragraph-level details.
Items from the middle and high school levels were drawn from the test split, while college-level items were taken from the validation and test splits, due to limited data in its test set.
Each MTF item was divided into four True/False/Not Given (TFNG) items, yielding triplets of (passage, statement, factuality label [\textit{True} or \textit{Not True}]).

Three experts with prior experience tutoring students for standardized English exams or authoring RC items independently identified the evidence sentences in the passage and labeled each item along the two cognitive dimensions.
For statements identified as \textit{False} within the \textit{Not True} cases, they produced minimally revised \textit{True} statements to enable annotation of transformation level. 
We retained only items where at least two annotators agreed on the same label; in cases of partial agreement, discrepancies were resolved through author adjudication.
The resulting annotated dataset, \dataName, is released for non-commercial research purposes under the RACE license. 
Further details on the annotation process and inter-annotator agreement are provided in Appendix~\ref{sec:appendix:annotation_detail}.

\subsection{Data Statistics}

Table~\ref{tab:data_statistic_test_demo} presents the statistics for the ReCo demonstration and test sets, and Table~\ref{tab:num_per_label} summarizes the distribution of items across labels.
For Evidence Scope, 50\% items are labeled as single-sentence evidence, while insufficient evidence items account for the lowest proportion.
For Transformation Level, items requiring inference are the most frequent, whereas transformed word matching items are the least common.
In the 3-level scheme, which includes multi-sentence comprehension items, inference items remain dominant.
These statistics reveal an imbalanced label distribution in our dataset, which is expected because the items are drawn from real exams, whose distributions may vary depending on the target proficiency level.
Further analysis of our dataset can be found in Appendix~\ref{sec:appendix:data_analysis}.

\begin{table}[t!]
\centering
\resizebox{0.9\columnwidth}{!}{
\begin{tabular}{lccc}
\toprule
\multirow{2}{*}{Split} & \multirow{2}{*}{\# Passage} & \multicolumn{2}{c}{\# Statement} \\ \cmidrule{3-4}
                                 &                            & \textit{True}            & \textit{Not True} \\ \midrule
\multirow{2}{*}{Test set}            & \multirow{2}{*}{151}       & 236 (409)       & 262           \\ \cmidrule{3-4}
                                 &                            & \multicolumn{2}{c}{498 (671)}   \\ \midrule
\multirow{2}{*}{Demonstration set}   & \multirow{2}{*}{83}        & 129 (222)       & 149           \\ \cmidrule{3-4}
                                 &                            & \multicolumn{2}{c}{278 (371)}   \\ \bottomrule
\end{tabular}}
\caption{\label{tab:data_statistic_test_demo} Number of passages and statements (\textit{True} / \textit{Not True} / Total) in the test and demonstration sets of ReCo. Parentheses indicate \textit{True} statements including annotator-revised versions of \textit{False} statement.}
\end{table}

\begin{table}[t]
\centering
\resizebox{\columnwidth}{!}{
\begin{tabular}{llll}
\toprule
\multicolumn{4}{c}{Evidence Scope}                                                      \\ \midrule
\multicolumn{3}{l}{Single-sentence Evidence}                & 388                  \\
\multicolumn{3}{l}{Multi-sentence Evidence}                 & 243                  \\
\multicolumn{3}{l}{Insufficient Evidence}                     & 145                  \\ \midrule
\multicolumn{4}{c}{Transformation Level}                                                \\ \midrule
\multicolumn{2}{c|}{\textit{5-level}}                & \multicolumn{2}{c}{\textit{3-level}}                  \\ \midrule
Word Matching             & \multicolumn{1}{l|}{73}  & \multirow{2}{*}{Word Matching} & \multirow{2}{*}{123} \\
Transformed Word Matching & \multicolumn{1}{l|}{36}  &                                &                      \\
Paraphrasing              & \multicolumn{1}{l|}{55}  & \multirow{2}{*}{Paraphrasing}  & \multirow{2}{*}{189} \\
Transformed Paraphrasing  & \multicolumn{1}{l|}{78}  &                                &                      \\
Inference                 & \multicolumn{1}{l|}{146} & Inference                      & 319                  \\ \bottomrule
\end{tabular}}
\caption{\label{tab:num_per_label} Distribution of examples in the \dataName\ dataset across Evidence Scope and Transformation Level. For the Transformation Level dimension, items labeled as insufficient evidence are excluded, and the 5-level scheme applies only to single-sentence comprehension items.}
\end{table}

\section{\label{experimental_setup}Experimental Setup}

We formulate the measurement of cognitive complexity as a classification task.
In the \textbf{Evidence Scope (ES) classification}, the model receives an instruction, a passage, a statement, and its factuality label (\textit{True} or \textit{Not True}), and predicts one of three evidence types: \textit{single}, \textit{multi}, or \textit{insufficient}. 
In the \textbf{Transformation Level (TL) classification}, applied to items with \textit{True} statements, the model estimates the degree of transformation using the task definition, passage, and statement.
We report performance using both the 5-level taxonomy---\textit{word matching (WM)}, \textit{transformed word matching (TWM)}, \textit{paraphrasing (P)}, \textit{transformed paraphrasing (TP)}, and \textit{inference (I)}---and a simplified 3-level version (\textit{WM}, \textit{P}, \textit{I}), which omits distinctions based on phrase reordering.
In the 3-level setting, predictions of TWM and TP are mapped to WM and P, respectively, and ground-truth labels for single-sentence evidence items are converted accordingly.
We report model performance using Macro F1; Micro F1 scores are provided in Appendix~\ref{sec:appendix:micro_performance}.

We evaluate eight  \textit{instruction-tuned} LLMs, including open-source models Gemma2-9B/27B ~\cite{gemma_2024}, Mistral-7B/24B~\cite{jiang2023mistral7b}, and Qwen2.5-7B/32B~\cite{qwen2.5}, as well as proprietary models GPT-4o and GPT-4o-mini~\cite{hurst2024gpt}.
Two prompting strategies are considered: Standard Prompting (SP), where the model receives a task definition and input and returns a label directly; and Chain-of-Thought Prompting (CoT), which encourages step-by-step reasoning before prediction~\cite{wei2022chain}.
Greedy decoding is used as the default inference method across all models and prompting strategies.
In the CoT setting, we additionally apply self-consistency decoding~\cite{wang2022self}, generating 10 samples with top-$k$=20, top-$p$=0.8, and temperature=0.7 and using priority answer.

Each strategy is evaluated under three prompting conditions—zero-shot, one-shot, and few-shot~\cite{brown2020language}—with the few-shot demonstrations covering items from all labels. 
Exemplars are sampled from the \dataName\ demonstration split and fixed across models to ensure consistency. 
To filter out overly trivial items that might inflate model performance, we exclude those that GPT-4o correctly classifies with a zero-shot CoT prompt.
Details on model versions and the experimental environment are provided in Appendix~\ref{sec:appendix:experimental_detail}.

\section{\label{sec:result} Results}

\begin{table}[t]
\centering
\small
\resizebox{\columnwidth}{!}{
\setlength{\tabcolsep}{0.98mm}
\begin{tabular}{lccccccccc}
\toprule
\multirow{2}{*}{Method}       & \multirow{2}{*}{\#Demo} & \multicolumn{2}{c}{Gemma2}  & \multicolumn{2}{c}{Mistral} & \multicolumn{2}{c}{Qwen2.5} & \multicolumn{2}{c}{GPT-4o}   \\
                              &                          & 9B        & 27B       & 7B        & 24B       & 7B        & 32B       & mini      & -         \\ \midrule
\multicolumn{10}{c}{\textit{Reading Comprehension}}                                                                                                                                              \\ \midrule
CoT  & 1                        & 78.6     & 82.4     & 59.8     & 74.2     & 74.8     & 82.8     & 80.6     & 84.4     \\ \midrule
\multicolumn{10}{c}{\textit{Evidence Scope Classification} [Human: 87.0]}                                                                                                                       \\ \midrule
\multirow{3}{*}{\begin{tabular}[c]{@{}l@{}}SP\end{tabular}}  & 0  & \underline{48.6} & 53.0     & 43.0     & 58.4     & 45.6     & 57.4     & 49.6     & 58.5     \\
                              & 1                        & 47.1     & \underline{56.6} & \underline{44.7} & \underline{60.2} & 47.7     & \underline{61.3} & \underline{52.7} & 62.0     \\
                              & 6                        & 48.5     & 51.7     & 39.4     & 56.8     & \underline{48.5} & 58.9     & 52.0     & \underline{65.6} \\ \midrule
\multirow{3}{*}{\begin{tabular}[c]{@{}l@{}}CoT\end{tabular}} & 0  & 53.6     & 58.3     & 15.9     & 62.9     & \underline{55.0} & 70.0     & 63.4     & 71.1     \\
                              & 1                        & \underline{61.3} & \underline{69.7} & \underline{53.1} & 65.9     & 53.0     & \underline{73.4} & \underline{66.7} & \textbf{\underline{74.8}} \\
                              & 6                        & 56.7     & 65.9     & 50.0     & \underline{66.4} & 53.8     & 70.5     & 63.0     & 68.4     \\ \midrule
\multirow{3}{*}{\begin{tabular}[c]{@{}l@{}}CoT (SC)\end{tabular}} & 0 & 55.5  & 63.4     & 12.7     & 67.2     & 55.6     & 73.1     & 65.1     & 71.8     \\
                              & 1                        & \textbf{\underline{64.2}} & \textbf{\underline{71.5}} & \textbf{\underline{58.0}} & \textbf{\underline{70.9}} & \textbf{\underline{57.5}} & \textbf{\underline{73.5}} & \textbf{\underline{71.0}} & \underline{73.2} \\
                              & 6                        & 58.3     & 68.1     & 49.6     & 65.9     & 51.7     & 71.0     & 66.5     & 72.3     \\ \midrule

\multicolumn{10}{c}{\textit{(3-level) Transformation Level Classification} [Human: 84.9]}                                                                                                      \\ \midrule
\multirow{3}{*}{\begin{tabular}[c]{@{}l@{}}SP\end{tabular}}  & 0  & \textbf{\underline{60.4}} & 50.8     & 46.9     & 72.9     & \underline{55.3} & 53.7     & 53.1     & \underline{62.3} \\
                              & 1                        & 54.2     & \underline{55.1} & \textbf{\underline{56.2}} & \underline{73.0} & 50.5     & \underline{56.5} & \underline{53.9} & 59.3     \\
                              & 8                        & 54.6     & 51.9     & 42.4     & 70.4     & 42.3     & 46.6     & 45.4     & 55.3     \\ \midrule
\multirow{3}{*}{\begin{tabular}[c]{@{}l@{}}CoT\end{tabular}} & 0  & \underline{52.9} & 51.4     & 43.3     & \underline{77.2} & 62.3     & 66.1     & 59.9     & \textbf{\underline{73.9}} \\
                              & 1                        & 49.9     & 53.7     & 44.9     & 70.2     & \textbf{\underline{69.4}} & \underline{68.6} & 62.3     & 72.9     \\
                              & 8                        & 50.9     & \underline{55.0} & \underline{46.6} & 71.3     & 56.8     & 67.2     & \textbf{\underline{65.1}} & 68.4     \\ \midrule
\multirow{3}{*}{\begin{tabular}[c]{@{}l@{}}CoT (SC)\end{tabular}} & 0 & \underline{58.5} & 55.9     & \underline{55.2} & \textbf{\underline{82.0}} & \underline{68.7} & \textbf{\underline{74.0}} & 58.4     & \underline{69.9} \\
                              & 1                        & 49.6     & 51.7     & 41.1     & 63.0     & 61.0     & 69.3     & \underline{60.3} & 68.0     \\
                              & 8                        & 57.2     & \textbf{\underline{56.3}} & 43.0     & 72.9     & 58.4     & 65.7     & 58.6     & 61.0     \\ \midrule

\multicolumn{10}{c}{\textit{(5-level) Transformation Level Classification} [Human: 83.0]}                                                                                                      \\ \midrule
\multirow{3}{*}{\begin{tabular}[c]{@{}l@{}}SP\end{tabular}}  & 0  & 32.2     & 27.4     & 22.1     & 45.6     & 31.5     & 32.1     & \underline{28.6} & 39.8     \\
                              & 1                        & \textbf{\underline{37.7}} & 30.7     & 23.2     & \underline{51.7} & \underline{33.0} & \underline{35.0} & 27.5     & \underline{40.5} \\
                              & 8                        & 30.2     & \underline{33.9} & \underline{26.8} & 47.5     & 24.4     & 28.4     & 27.6     & 37.7     \\ \midrule
\multirow{3}{*}{\begin{tabular}[c]{@{}l@{}}CoT\end{tabular}} & 0  & 32.1     & 32.9     & 24.4     & \underline{55.0} & 35.1     & \underline{52.2} & 43.5     & \textbf{\underline{61.3}} \\
                              & 1                        & 32.0     & 32.8     & 26.3     & 42.6     & \underline{42.7} & 45.4     & \textbf{\underline{44.5}} & 60.0     \\
                              & 8                        & \underline{35.7} & \underline{34.0} & \underline{27.6} & 45.4     & 39.6     & 43.4     & 43.3     & 49.6     \\ \midrule
\multirow{3}{*}{\begin{tabular}[c]{@{}l@{}}CoT (SC)\end{tabular}} & 0 & 33.8     & \textbf{\underline{38.9}} & 26.6     & \textbf{\underline{58.7}} & 31.9     & \textbf{\underline{56.0}} & \underline{42.1} & 53.1     \\
                              & 1                        & 35.3     & 35.9     & \textbf{\underline{28.0}} & 38.4     & \textbf{\underline{44.1}} & 48.2     & 40.0     & \underline{54.6} \\
                              & 8                        & \underline{36.9} & 32.1     & 26.4     & 48.3     & 35.6     & 43.4     & 37.5     & 43.4     \\ \bottomrule
\end{tabular}}\caption{\label{tab:main_result} 
Performance of LLMs on the RC task and the ES and TL classification tasks.
Greedy decoding is the default inference method, and SC indicates that self-consistency decoding is used.
\textbf{Bolded} values denote each model's best score per task; \underline{underlined} values indicate the best score within each method group across demonstration settings.}
\end{table}

Table~\ref{tab:main_result} presents the performance of LLMs and human experts on the ES and TL classification tasks.
Human performance is computed using annotators’ initial labels, before applying inter-annotator agreement filtering and adjudication.
The table also reports LLMs’ performance on the RC task, which requires the model to determine whether a statement is true or not based on the passage.

Before evaluating LLMs’ ability to analyze the cognitive complexity, we first examined their performance on the RC task itself.
According to the results, all models except the Mistral family achieved Macro F1 scores above 74.0, with larger models approaching 85.0, indicating strong reading comprehension ability.
This result confirms that the RC items in \dataName\ are generally easy for current LLMs, and thus that errors in cognitive complexity prediction are unlikely to stem from failures in basic comprehension or answer reasoning.
However, for smaller models such as Mistral-7B, incomplete comprehension may still contribute to some degree of performance variation.

In the ES classification task, GPT-4o achieves the highest score (74.8) with a one-shot CoT prompt, and Qwen2.5-32B performs comparably (73.5).
Yet all models fall short of expert performance (87.0), highlighting the difficulty of modeling human cognitive processes in evidence selection.
Within open-source model families, larger models tend to yield comparable performance across variants---particularly under self-consistency decoding---, while smaller models exhibit more divergent results.
Mistral-7B, in particular, exhibits unstable behavior, occasionally yielding unexpectedly low scores despite identical prompts.
As expected, zero-shot prompting underperforms compared to demonstration-based settings; however, in large models, few-shot prompting occasionally led to performance degradation.

In the TL classification task with the 3-level taxonomy, Mistral-24B (82.0) and Qwen2.5-32B (74.0) outperform GPT-4o (73.9), approaching human performance (84.9).
Under the 5-level taxonomy, however, overall model performance decreases, underscoring the challenge of capturing phrase reordering between statements and evidence.
Nevertheless, open-source models again achieve performance comparable to GPT-4o.
By contrast, Gemma2-27B consistently underperforms, trailing even smaller models, which suggests limitations in handling lexical and syntactic transformations.
Prompting effects are less consistent than in ES classification: optimal configurations vary by model, and in some cases zero-shot prompting even outperforms demonstration-based prompting.
Self-consistency decoding improves results in several settings, particularly zero-shot CoT, but its gains are less stable than in ES classification.

\begin{table*}[t]
\centering
\small
\begin{tabular}{lrrrrrrrr}
\toprule
\multirow{2}{*}{Subtask}            & \multicolumn{2}{c}{Gemma2}                       & \multicolumn{2}{c}{Mistral}                      & \multicolumn{2}{c}{Qwen2.5}                     & \multicolumn{2}{c}{GPT-4o}                       \\
                                 & \multicolumn{1}{c}{9B} & \multicolumn{1}{c}{27B} & \multicolumn{1}{c}{7B} & \multicolumn{1}{c}{24B} & \multicolumn{1}{c}{7B} & \multicolumn{1}{c}{32B} & \multicolumn{1}{c}{mini} & \multicolumn{1}{c}{-} \\ \midrule
1.1. Falsifiability Judgment     & 80.6                   & 77.0                   & 65.9                  & 80.8                   & 76.0                  & \textbf{88.6}          & 77.0                     & 85.5                 \\
1.2. Evidence Sentence Counting  & 46.4                   & \textbf{47.7}          & 41.1                  & 44.0                   & 43.7                  & 47.6                   & 41.0                     & 44.6                 \\ \midrule
2.1. Inference Detection         & 72.9                   & 75.3                   & 65.4                  & \textbf{82.4}          & 66.6                  & 81.6                   & 81.4                     & 79.8                 \\
2.2. Paraphrasing Detection      & 81.3                   & 84.3                   & 66.6                  & 84.9                   & 72.0                  & 85.3                   & 87.4                     & \textbf{88.5}        \\
2.3. Phrase Reordering Detection & 58.0                   & \textbf{67.9}          & 58.6                  & 61.0                   & 41.3                  & 47.5                   & 66.6                     & 63.3                 \\ \bottomrule
\end{tabular}
\caption{\label{tab:subtask_result} LLM performance on subtasks measuring fine-grained abilities required for the main classification tasks. \textbf{Bolded} values denote the best model for each subtask.}
\end{table*}

In summary, \textbf{while LLMs do not fully align with expert judgments, they demonstrate strong potential as estimators of cognitive complexity in RC items, especially in ES and 3-level TL classifications.} 
Notably, the competitive performance of open-source models relative to GPT-4o suggests that reliance on proprietary LLMs may not be necessary for analyzing cognitive complexity.
While we employ simple prompting methods to better isolate inherent model capabilities, more advanced prompt engineering may yield further improvements.

\section{Analysis}
\subsection{Fine-Grained Feature Analysis}

To better understand the capabilities and limitations of LLMs, we decomposed the two classification tasks into a set of fine-grained subtasks.
For each subtask, we constructed few-shot CoT prompts by adapting those used in the main experiments, with instructions tailored to the specific cognitive feature under evaluation.
Results are reported in Table~\ref{tab:subtask_result}.

The ES classification task was divided into two core subtasks:
\textbf{Subtask 1.1: Falsifiability Judgment} -- determining whether a \textit{Not True} statement is \textit{False} (contradicted) or \textit{Not Given} (lacking sufficient evidence).
\textbf{Subtask 1.2: Evidence Sentence Counting} -- identifying how many sentences are required to support or refute a statement.
Most models performed reliably on falsifiability classification, but performance was substantially lower for evidence sentence counting.
This suggests that while LLMs can distinguish between refuted and unsupported statements, they struggle to accurately identify the full set of sentences referenced by human annotators when solving the item.

The TL classification task was evaluated through a hierarchical breakdown of transformation types, allowing for a more detailed examination of how LLMs handle different forms of linguistic transformation:
\textbf{Subtask 2.1. Inference Detection} -- distinguishing inference-based statements from those explainable by surface-level transformations, such as paraphrasing or phrase reordering.
\textbf{Subtask 2.2. Paraphrasing Detection} -- identifying whether a statement is a lexical rephrasing or a verbatim restatement.
\textbf{Subtask 2.3: Phrase Reordering Detection} -- detecting reordering of words and phrases.
According to the results, model performance showed distinct strengths and weaknesses across subtasks.
GPT-4o-mini performed comparably to, or even better than, GPT-4o---especially on inference and phrase reordering detection, where GPT-4o underperformed.
Paraphrasing detection was handled well across models, while phrase reordering remained especially challenging.

Overall, LLMs showed consistent performance on falsifiability classification (Subtask 1.1) and paraphrase detection (Subtask 2.2), but struggled with evidence sentence counting (Subtask 1.2) and phrase reordering detection (Subtask 2.3).
While models demonstrated stable answer prediction performance (as shown in Table~\ref{tab:main_result}), our analysis indicates that even when they solved items correctly, they often failed to explicitly capture the cognitive features underlying their reasoning process, pointing to a limitation in metacognitive awareness.

\subsection{Error Analysis}

\begin{figure}[t]
\centering
\includegraphics[width=0.7\columnwidth]{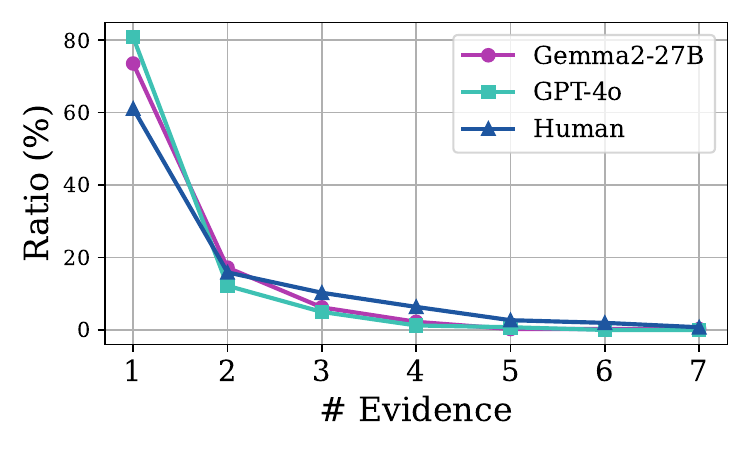}
\caption{\label{fig:evidence_sentence_distribution} Distribution of the number of evidence sentences selected by LLMs and humans.}
\end{figure}

We further examined the prediction tendencies of GPT-4o and the best-performing model in each task.
Figure~\ref{fig:evidence_sentence_distribution} shows the distribution of the number of evidence sentences predicted by Gemma2-27B and GPT-4o, compared to human-selected evidence.
LLMs tend to select fewer sentences than humans, often defaulting to a single sentence as evidence.
Table~\ref{tab:evidence_sentence_selection} reports the degree of alignment between the evidence sentences selected by each model and those identified by experts.
As in the previous observation, recall scores were consistently lower than corresponding precision scores, indicating that models often fail to explicitly retrieve all the sentences required by humans to solve the item.
One possible explanation is that LLMs—by encoding the entire passage before engaging in step-by-step reasoning—tend to assign insufficient attention to sentences containing seemingly minor details that nevertheless exert a significant influence on the answer choice.

\begin{table}[t]
\centering
\small
\begin{tabular}{lccc}
\toprule
Evidence Selection & Precision & Recall & F1    \\ \midrule
Gemma2-27B         & 86.4      & 78.3   & 78.8  \\
Mistral-24B        & 82.4      & 74.4   & 74.3  \\
Qwen2.5-32B        & 85.4      & 76.9   & 77.4  \\
GPT-4o             & 88.8      & 79.2   & 80.0  \\ \bottomrule
\end{tabular}
\caption{\label{tab:evidence_sentence_selection} Model performance on evidence sentence selection. For each instance, precision, recall, and F1 were computed by comparing the predicted and reference evidence sets over their union, and the final scores were obtained by averaging these instance-level values across the dataset.}
\end{table}

\begin{figure}[t]
\centering
\includegraphics[width=\columnwidth]{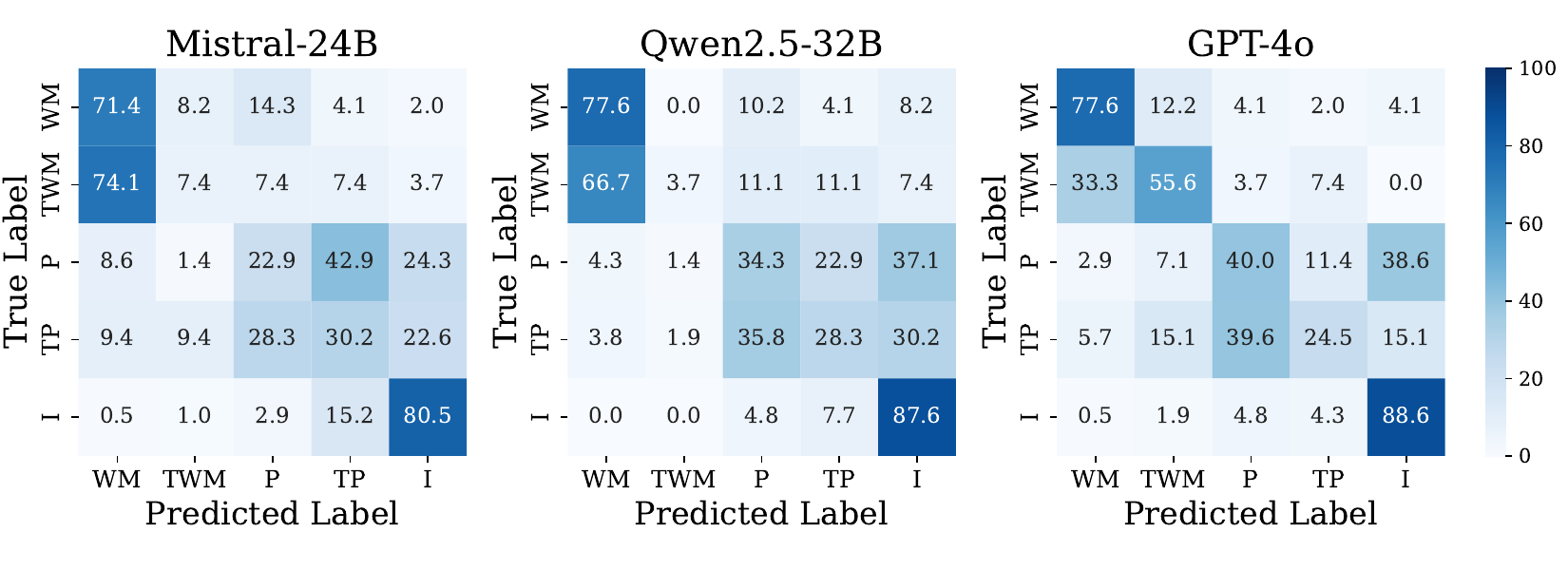}
\caption{\label{fig:tl_label_distribution} Distribution of TL labels predicted by LLMs for single-sentence evidence items.}
\end{figure}

Figure~\ref{fig:tl_label_distribution} presents the distribution of TL predictions for items with \textit{single-sentence evidence}.
LLMs consistently achieved higher precision for \textit{WM} and \textit{I} items than for other categories.
However, they frequently misclassified \textit{TWM} as \textit{WM}, failing to detect subtle reordering of phrases.
Models also showed confusion between \textit{P} and \textit{TP}, often misclassifying them as each other or as \textit{I}.
These patterns align with the earlier results in Table~\ref{tab:subtask_result}, particularly for phrase reordering detection, where all models struggled.
Representative error cases illustrating these confusions are analyzed in the case study.

\begin{figure*}[t]
\centering
\includegraphics[width=\textwidth]{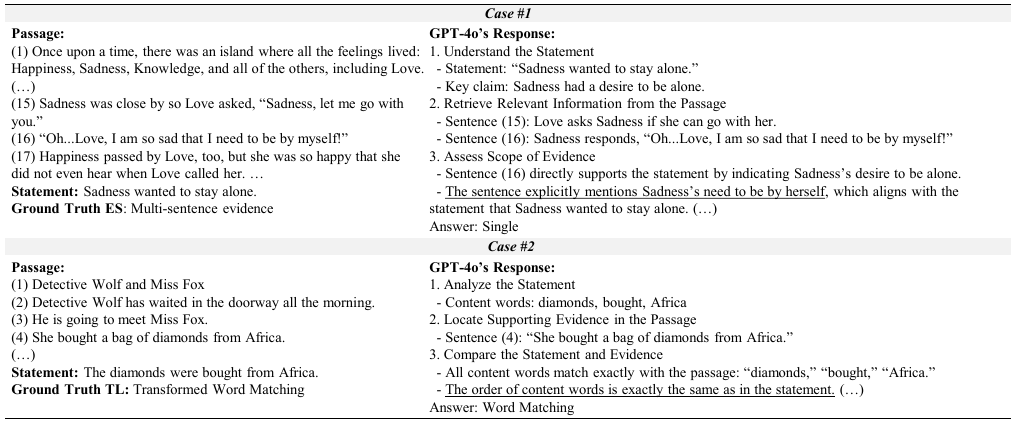}
\caption{\label{fig:error_analysis} Representative error cases with GPT-4o’s responses.}
\end{figure*}

\subsection{\label{sec:more_exp} Evaluating the Effect of Deep Reasoning on Cognitive Complexity Estimation}

We investigated whether LLMs with advanced reasoning capabilities are also effective at analyzing cognitive complexity.
For this experiment, we used Qwen3-32B~\cite{qwen3technicalreport}, a reasoning-specialized model that supports a ``thinking mode'' designed for deep reasoning and self-reflection.
We compared the performance of Qwen3-32B in two settings---with and without thinking mode---on the cognitive difficulty classification tasks.
As a baseline, we also included Qwen2.5-32B, a model from the same family but not specialized for reasoning.

In Table~\ref{tab:qwen3}, we observed that Qwen3-32B in thinking mode underperformed its non-thinking-mode counterpart, achieving lower F1 scores across both classification tasks.
These results suggest that advanced reasoning and self-reflection capabilities do not necessarily enhance a model's ability to classify cognitive complexity features such as ES and TL.
This may be because these tasks do not require complex multi-step reasoning, but rather fine-grained categorization of human cognitive processes---something better handled through intuitive pattern recognition than abstract reasoning.
The experimental results align with this interpretation and highlight the distinction between deep reasoning ability and the metacognitive awareness.

\begin{table}[t]
\centering
\small
\begin{tabular}{lccc}
\toprule
Model          & ES             & \begin{tabular}[c]{@{}c@{}}TL\\(\textit{5-level})\end{tabular}   & \begin{tabular}[c]{@{}c@{}}TL\\(\textit{3-level})\end{tabular} \\ \midrule
Qwen2.5-32B    & 70.0           & 52.2       & 66.1 \\
Qwen3-32B$_{non-thinking}$      & 65.4        & 58.6       & 78.2 \\
Qwen3-32B$_{thinking}$   & 62.6        & 56.3      & 69.4  \\ \bottomrule
\end{tabular}
\caption{\label{tab:qwen3} Performance comparison of LLMs with and without deep-thinking capabilities.}
\end{table}

\subsection{Case Study}

Figure~\ref{fig:error_analysis} presents error cases that illustrate common failure patterns in LLMs.
We analyze two examples where GPT-4o incorrectly classifies items in the ES and TL classification tasks.
In Case \#1, the model misclassifies an item requiring \textit{multi-sentence evidence} as \textit{single-sentence evidence}.
The passage involves multiple characters in dialogue, and correctly identifying the speaker of a specific utterance is essential for determining the truth of the statement.
According to annotators, both Sentence (15) and Sentence (16) are required: Sentence (15) establishes that the subject is ``Sadness,'' while Sentence (16) describes her action.
However, the model references Sentence (15) in its explanation and correctly links ``Sadness'' to the action in Sentence (16), yet asserts: ``Sentence (16) explicitly mentions Sadness’s need to be by herself.''
The case reveals a metacognitive failure: the model arrives at the correct factual judgment but fails to recognize the reasoning behavior it engaged in to reach that conclusion.

In Case \#2, the item labeled as \textit{TWM} was incorrectly classified as \textit{WM}.
A comparison with Sentence (4) reveals a reordering of content words caused by a shift to the passive construction.
However, GPT-4o claims: “The order of content words in the evidence is exactly the same as in the statement.”
This suggests that the model sometimes fails to detect subtle syntactic transformations.
These two cases together suggest that, even when LLMs provide seemingly coherent explanations, they may miss structural cues and fail to reflect on the reasoning processes underlying their own answers.
Such blind spots underscore a persistent challenge of detecting and modeling the cognitive features underlying human problem-solving processes.

\section{Conclusion}

The cognitive complexity of the problem-solving process is a key factor for analyzing the prior difficulty of RC items, yet no scalable method currently exists for automatically measuring it.
In this study, we investigated whether LLMs can predict the cognitive complexity of RC items through two cognitively grounded dimensions: Evidence Scope and Transformation Level.
To this end, we constructed \dataName---a benchmark dataset of RC items annotated along these two dimensions---and conducted a comprehensive evaluation of eight LLMs under diverse prompting and decoding configurations.

The results show that LLMs have strong potential as proxies for cognitive complexity estimation, with some open-source models—such as Qwen2.5-32B—achieving performance comparable to proprietary systems like GPT-4o.
Nevertheless, LLMs are not fully aligned with human experts and exhibit limitations in their metacognitive awareness---particularly in detecting phrase reordering or identifying all necessary evidence from the passage.
We hope our findings encourage further research in item difficulty estimation and difficulty-controlled item generation---contributing to the development of more interpretable and cognitively aligned educational NLP systems.

\section*{Limitations}

\paragraph{Label Imbalance and Data Scale.}
Our dataset exhibits label imbalance across cognitive dimensions, which is an inevitable outcome of annotating RC items randomly sampled from assessments.
The scarcity of certain labels can be attributed to several factors: such item types may be more difficult to create, less emphasized in instructional practice, or underrepresented in the specific assessments from which our samples were drawn.
However, supplementing these underrepresented categories would require additional large-scale annotation, which was infeasible under our budget constraints.

Moreover, constructing the dataset required costly expert annotation—three raters per item—which constrained our ability to perform supervised fine-tuning.
Nevertheless, our dataset comprises 776 items, a substantially larger resource than the previously released CMCQRD dataset containing only 289 items~\cite{mullooly2023cambridge}.
Our findings also suggest that larger models (24B–32B) exhibit more robust in-context classification than smaller ones (7B–9B), thus highlighting opportunities for future work on data augmentation and knowledge distillation to build smaller yet effective models.

\paragraph{Limited Coverage of RC Item Types.}

The two cognitive dimensions investigated—Evidence Scope and Transformation Level—do not generalize across all RC item types in capturing their cognitive complexity.
For instance, questions targeting main ideas or author intent inherently require multi-sentence inference.
In contrast, the dimensions we employ are most relevant to factual detail questions such as TFNG, MTF, and WH-questions.
Because different RC item types involve distinct cognitive complexity factors, addressing all of them comprehensively would be beyond the scope of a single study.
In this work, we therefore focus on two key dimensions best captured by factual detail questions, as this question format spans a wide range of cognitive complexity—from single-sentence word matching to multi-sentence inference.
Our goal is to establish a foundation for future research on additional factors across a broader set of RC item types.

\paragraph{Limits of a Single Factor in Explaining Item Difficulty.}
Finally, the two dimensions studied here represent only a subset of the many factors influencing RC item difficulty. 
Establishing a direct or linear relationship between a single factor and difficulty would require controlling for all other variables, which is not the case in our dataset.
For instance, a multi-sentence inference item may not necessarily be harder than a single-sentence paraphrasing item if the latter references a more complex passage.
Thus, \textit{these two factors alone cannot fully account for item difficulty}.
This study instead focuses on testing whether LLMs can estimate cognitively grounded dimensions that have traditionally relied on human annotation.

\paragraph{Potential Data Contamination}
We constructed \dataName\ exclusively from the test split of RACE++, with college-level items supplemented from the validation split to meet the required sample size.
Nevertheless, we acknowledge that some LLMs may have incorporated the RACE++ test and validation splits into their pre-training corpora.
Even so, we believe the impact on our evaluation is limited: while exposure to RACE++ in its original multiple-choice format may help models recall specific answers, it does not directly expose the cognitive-complexity labels required in \dataName.
Consistent with this, our results show that high RC performance does not translate into accurate cognitive complexity classification, suggesting that incidental memorization of RACE++ items does not confer a meaningful advantage on our tasks.

\section*{Acknowledgments}
This research was supported by Culture, Sports and Tourism R\&D Program through the Korea Creative Content Agency grant funded by the Ministry of Culture, Sports and Tourism in 2025 (Project Name: Development of an AI-Based Korean Diagnostic System for Efficient Korean Speaking Learning by Foreigners, Project Number: RS-2025-02413038, Contribution Rate: 45\%); by the IITP (Institute of Information \& Coummunications Technology Planning \& Evaluation) - ITRC (Information Technology Research Center) grant funded by the Korea government (Ministry of Science and ICT) (IITP-2026-RS-2024-00437866, Contribution Rate: 45\%); and by Institute of Information \& communications Technology Planning \& Evaluation (IITP) grant funded by the Korea government (MSIT) (No.RS-2019-II191906, Artificial Intelligence Graduate School Program (POSTECH), Contribution Rate: 10\%).
We also thank Jonghwi Kim for valuable feedback on this paper.

\bibliography{custom}

\begin{thebibliography}{49}
\providecommand{\natexlab}[1]{#1}

\bibitem[{AlKhuzaey et~al.(2024)AlKhuzaey, Grasso, Payne, and Tamma}]{alkhuzaey2024text}
Samah AlKhuzaey, Floriana Grasso, Terry~R Payne, and Valentina Tamma. 2024.
\newblock Text-based question difficulty prediction: A systematic review of automatic approaches.
\newblock \emph{International Journal of Artificial Intelligence in Education}, 34(3):862--914.

\bibitem[{Anderson(1972)}]{anderson1972construct}
Richard~C Anderson. 1972.
\newblock How to construct achievement tests to assess comprehension.
\newblock \emph{Review of educational research}, 42(2):145--170.

\bibitem[{Benedetto et~al.(2021)Benedetto, Aradelli, Cremonesi, Cappelli, Giussani, and Turrin}]{benedetto2021application}
Luca Benedetto, Giovanni Aradelli, Paolo Cremonesi, Andrea Cappelli, Andrea Giussani, and Roberto Turrin. 2021.
\newblock On the application of transformers for estimating the difficulty of multiple-choice questions from text.
\newblock In \emph{Proceedings of the 16th workshop on innovative use of NLP for building educational applications}, pages 147--157.

\bibitem[{Bloom et~al.(1956)Bloom, Engelhart, Furst, Hill, Krathwohl et~al.}]{bloom1956taxonomy}
Benjamin~S Bloom, Max~D Engelhart, Edward~J Furst, Walker~H Hill, David~R Krathwohl, et~al. 1956.
\newblock \emph{Taxonomy of educational objectives: The classification of educational goals. Handbook 1: Cognitive domain}.
\newblock Longman New York.

\bibitem[{Bormuth et~al.(1970)Bormuth, Manning, Carr, and Pearson}]{bormuth1970children}
John~R Bormuth, John Manning, Julian Carr, and David Pearson. 1970.
\newblock Children's comprehension of between-and within-sentence syntactic structures.
\newblock \emph{Journal of educational psychology}, 61(5):349.

\bibitem[{Brown et~al.(2020)Brown, Mann, Ryder, Subbiah, Kaplan, Dhariwal, Neelakantan, Shyam, Sastry, Askell et~al.}]{brown2020language}
Tom Brown, Benjamin Mann, Nick Ryder, Melanie Subbiah, Jared~D Kaplan, Prafulla Dhariwal, Arvind Neelakantan, Pranav Shyam, Girish Sastry, Amanda Askell, et~al. 2020.
\newblock Language models are few-shot learners.
\newblock \emph{Advances in neural information processing systems}, 33:1877--1901.

\bibitem[{Byrd and Srivastava(2022)}]{byrd2022predicting}
Matthew Byrd and Shashank Srivastava. 2022.
\newblock Predicting difficulty and discrimination of natural language questions.
\newblock In \emph{Proceedings of the 60th Annual Meeting of the Association for Computational Linguistics (Volume 2: Short Papers)}, pages 119--130.

\bibitem[{Choi and Moon(2020)}]{choi2020predicting}
Inn-Chull Choi and Youngsun Moon. 2020.
\newblock Predicting the difficulty of efl tests based on corpus linguistic features and expert judgment.
\newblock \emph{Language Assessment Quarterly}, 17(1):18--42.

\bibitem[{Clark et~al.(2019)Clark, Lee, Chang, Kwiatkowski, Collins, and Toutanova}]{clark2019boolq}
Christopher Clark, Kenton Lee, Ming-Wei Chang, Tom Kwiatkowski, Michael Collins, and Kristina Toutanova. 2019.
\newblock Boolq: Exploring the surprising difficulty of natural yes/no questions.
\newblock In \emph{Proceedings of the 2019 Conference of the North American Chapter of the Association for Computational Linguistics: Human Language Technologies, Volume 1 (Long and Short Papers)}, pages 2924--2936.

\bibitem[{Devlin et~al.(2019)Devlin, Chang, Lee, and Toutanova}]{devlin2019bert}
Jacob Devlin, Ming-Wei Chang, Kenton Lee, and Kristina Toutanova. 2019.
\newblock Bert: Pre-training of deep bidirectional transformers for language understanding.
\newblock In \emph{Proceedings of the 2019 conference of the North American chapter of the association for computational linguistics: human language technologies, volume 1 (long and short papers)}, pages 4171--4186.

\bibitem[{Dutulescu et~al.(2024)Dutulescu, Ruseti, Dascalu, and Mcnamara}]{dutulescu2024hard}
Andreea Dutulescu, Stefan Ruseti, Mihai Dascalu, and Danielle Mcnamara. 2024.
\newblock How hard can this question be? an exploratory analysis of features assessing question difficulty using llms.
\newblock In \emph{Proceedings of the 17th International Conference on Educational Data Mining}, pages 802--808.

\bibitem[{Embretson and Wetzel(1987)}]{embretson1987component}
Susan~E Embretson and C~Douglas Wetzel. 1987.
\newblock Component latent trait models for paragraph comprehension tests.
\newblock \emph{Applied psychological measurement}, 11(2):175--193.

\bibitem[{Freedle and Kostin(1991)}]{freedle1991prediction}
Roy Freedle and Irene Kostin. 1991.
\newblock The prediction of gre reading comprehension item difficulty for expository prose passages for each of three item types: Main ideas, inferences and explicit statements.
\newblock \emph{ETS Research Report Series}, 1991(2):i--53.

\bibitem[{Hambleton and Jones(1993)}]{hambleton1993comparison}
Ronald~K. Hambleton and Ronald~W. Jones. 1993.
\newblock \href {https://doi.org/10.1111/j.1745-3992.1993.tb00543.x} {Comparison of classical test theory and item response theory and their applications to test development}.
\newblock \emph{Educational Measurement: Issues and Practice}, 12(3):38--47.

\bibitem[{Hsu et~al.(2018)Hsu, Lee, Chang, and Sung}]{hsu2018automated}
Fu-Yuan Hsu, Hahn-Ming Lee, Tao-Hsing Chang, and Yao-Ting Sung. 2018.
\newblock Automated estimation of item difficulty for multiple-choice tests: An application of word embedding techniques.
\newblock \emph{Information Processing \& Management}, 54(6):969--984.

\bibitem[{Huang et~al.(2017)Huang, Liu, Chen, Zhao, Gao, Wei, Su, and Hu}]{huang2017question}
Zhenya Huang, Qi~Liu, Enhong Chen, Hongke Zhao, Mingyong Gao, Si~Wei, Yu~Su, and Guoping Hu. 2017.
\newblock Question difficulty prediction for reading problems in standard tests.
\newblock In \emph{Proceedings of the AAAI conference on artificial intelligence}, volume~31.

\bibitem[{Hurst et~al.(2024)Hurst, Lerer, Goucher, Perelman, Ramesh, Clark, Ostrow, Welihinda, Hayes, Radford et~al.}]{hurst2024gpt}
Aaron Hurst, Adam Lerer, Adam~P Goucher, Adam Perelman, Aditya Ramesh, Aidan Clark, AJ~Ostrow, Akila Welihinda, Alan Hayes, Alec Radford, et~al. 2024.
\newblock Gpt-4o system card.
\newblock \emph{arXiv preprint arXiv:2410.21276}.

\bibitem[{Hutzler et~al.(2014)Hutzler, David, Avigal, and Azoulay}]{hutzler2014learning}
Dorit Hutzler, Esther David, Mireille Avigal, and Rina Azoulay. 2014.
\newblock Learning methods for rating the difficulty of reading comprehension questions.
\newblock In \emph{2014 ieee international conference on software science, technology and engineering}, pages 54--62. IEEE.

\bibitem[{Jain et~al.(2025)Jain, Hollander, He, Tang, Zhang, and Sabatini}]{jain2025exploring}
Yoshee Jain, John Hollander, Amber He, Sunny Tang, Liang Zhang, and John Sabatini. 2025.
\newblock Exploring the potential of large language models for estimating the reading comprehension question difficulty.
\newblock In \emph{International Conference on Human-Computer Interaction}, pages 202--213. Springer.

\bibitem[{Jiang et~al.(2023)Jiang, Sablayrolles, Mensch, Bamford, Chaplot, de~las Casas, Bressand, Lengyel, Lample, Saulnier, Lavaud, Lachaux, Stock, Scao, Lavril, Wang, Lacroix, and Sayed}]{jiang2023mistral7b}
Albert~Q. Jiang, Alexandre Sablayrolles, Arthur Mensch, Chris Bamford, Devendra~Singh Chaplot, Diego de~las Casas, Florian Bressand, Gianna Lengyel, Guillaume Lample, Lucile Saulnier, Lélio~Renard Lavaud, Marie-Anne Lachaux, Pierre Stock, Teven~Le Scao, Thibaut Lavril, Thomas Wang, Timothée Lacroix, and William~El Sayed. 2023.
\newblock \href {https://arxiv.org/abs/2310.06825} {Mistral 7b}.
\newblock \emph{Preprint}, arXiv:2310.06825.

\bibitem[{Kapoor et~al.(2025)Kapoor, Truong, Haber, Ruiz-Primo, and Domingue}]{kapoor2025prediction}
Radhika Kapoor, Sang~T Truong, Nick Haber, Maria~Araceli Ruiz-Primo, and Benjamin~W Domingue. 2025.
\newblock Prediction of item difficulty for reading comprehension items by creation of annotated item repository.
\newblock \emph{arXiv preprint arXiv:2502.20663}.

\bibitem[{Ko et~al.(2024)Ko, Park, Park, and Seo}]{ko2024hierarchical}
Miyoung Ko, Sue Park, Joonsuk Park, and Minjoon Seo. 2024.
\newblock Hierarchical deconstruction of llm reasoning: A graph-based framework for analyzing knowledge utilization.
\newblock In \emph{Proceedings of the 2024 Conference on Empirical Methods in Natural Language Processing}, pages 4995--5027.

\bibitem[{Lai et~al.(2017)Lai, Xie, Liu, Yang, and Hovy}]{lai2017race}
Guokun Lai, Qizhe Xie, Hanxiao Liu, Yiming Yang, and Eduard Hovy. 2017.
\newblock Race: Large-scale reading comprehension dataset from examinations.
\newblock In \emph{Proceedings of the 2017 Conference on Empirical Methods in Natural Language Processing}, pages 785--794.

\bibitem[{Lalor et~al.(2019)Lalor, Wu, and Yu}]{lalor2019learning}
John~P Lalor, Hao Wu, and Hong Yu. 2019.
\newblock Learning latent parameters without human response patterns: Item response theory with artificial crowds.
\newblock In \emph{Proceedings of the 2019 Conference on Empirical Methods in Natural Language Processing and the 9th International Joint Conference on Natural Language Processing (EMNLP-IJCNLP)}, pages 4249--4259.

\bibitem[{Liang et~al.(2019)Liang, Li, and Yin}]{liang2019new}
Yichan Liang, Jianheng Li, and Jian Yin. 2019.
\newblock A new multi-choice reading comprehension dataset for curriculum learning.
\newblock In \emph{Asian Conference on Machine Learning}, pages 742--757. PMLR.

\bibitem[{Liusie et~al.(2023)Liusie, Raina, Mullooly, Knill, and Gales}]{liusie2023analysis}
Adian Liusie, Vatsal Raina, Andrew Mullooly, Kate Knill, and Mark~JF Gales. 2023.
\newblock Analysis of the cambridge multiple-choice questions reading dataset with a focus on candidate response distribution.
\newblock \emph{arXiv preprint arXiv:2306.13047}.

\bibitem[{Loper and Bird(2002)}]{loper2002nltk}
Edward Loper and Steven Bird. 2002.
\newblock Nltk: The natural language toolkit.
\newblock \emph{arXiv preprint cs/0205028}.

\bibitem[{Lord(1980)}]{lord1980irt}
Frederic~M. Lord. 1980.
\newblock \href {https://doi.org/10.4324/9780203056615} {\emph{Applications of Item Response Theory To Practical Testing Problems}}.
\newblock Routledge.

\bibitem[{Lu and Wang(2024)}]{lu2024generative}
Xinyi Lu and Xu~Wang. 2024.
\newblock Generative students: Using llm-simulated student profiles to support question item evaluation.
\newblock In \emph{Proceedings of the Eleventh ACM Conference on Learning@ Scale}, pages 16--27.

\bibitem[{McNamara et~al.(2014)McNamara, Graesser, McCarthy, and Cai}]{mcnamara2014automated}
Danielle~S McNamara, Arthur~C Graesser, Philip~M McCarthy, and Zhiqiang Cai. 2014.
\newblock \emph{Automated evaluation of text and discourse with Coh-Metrix}.
\newblock Cambridge University Press.

\bibitem[{Mikolov et~al.(2013)Mikolov, Chen, Corrado, and Dean}]{mikolov2013efficient}
Tomas Mikolov, Kai Chen, Greg Corrado, and Jeffrey Dean. 2013.
\newblock Efficient estimation of word representations in vector space.
\newblock \emph{arXiv preprint arXiv:1301.3781}.

\bibitem[{Mullooly et~al.(2023)Mullooly, Andersen, Benedetto, Buttery, Caines, Gales, Karatay, Knill, Liusie, Raina et~al.}]{mullooly2023cambridge}
Andrew Mullooly, {\O}istein Andersen, Luca Benedetto, Paula Buttery, Andrew Caines, Mark~JF Gales, Yasin Karatay, Kate Knill, Adian Liusie, Vatsal Raina, et~al. 2023.
\newblock The cambridge multiple-choice questions reading dataset.

\bibitem[{Pandarova et~al.(2019)Pandarova, Schmidt, Hartig, Boubekki, Jones, and Brefeld}]{pandarova2019predicting}
Irina Pandarova, Torben Schmidt, Johannes Hartig, Ahc{\`e}ne Boubekki, Roger~Dale Jones, and Ulf Brefeld. 2019.
\newblock Predicting the difficulty of exercise items for dynamic difficulty adaptation in adaptive language tutoring.
\newblock \emph{International Journal of Artificial Intelligence in Education}, 29(3):342--367.

\bibitem[{Park(2004)}]{park2004comparison}
Gi-Pyo Park. 2004.
\newblock Comparison of l2 listening and reading comprehension by university students learning english in korea.
\newblock \emph{Foreign Language Annals}, 37(3):448--458.

\bibitem[{Park et~al.(2024)Park, Park, Won, and Kim}]{park2024large}
Jae-Woo Park, Seong-Jin Park, Hyun-Sik Won, and Kang-Min Kim. 2024.
\newblock Large language models are students at various levels: Zero-shot question difficulty estimation.
\newblock In \emph{Findings of the Association for Computational Linguistics: EMNLP 2024}, pages 8157--8177.

\bibitem[{Rafatbakhsh and Ahmadi(2023)}]{rafatbakhsh2023predicting}
Elaheh Rafatbakhsh and Alireza Ahmadi. 2023.
\newblock Predicting the difficulty of efl reading comprehension tests based on linguistic indices.
\newblock \emph{Asian-Pacific Journal of Second and Foreign Language Education}, 8(1):41.

\bibitem[{Raina and Gales(2024)}]{raina2024question}
Vatsal Raina and Mark Gales. 2024.
\newblock Question difficulty ranking for multiple-choice reading comprehension.
\newblock \emph{arXiv preprint arXiv:2404.10704}.

\bibitem[{Rajpurkar et~al.(2016)Rajpurkar, Zhang, Lopyrev, and Liang}]{rajpurkar2016squad}
Pranav Rajpurkar, Jian Zhang, Konstantin Lopyrev, and Percy Liang. 2016.
\newblock Squad: 100,000+ questions for machine comprehension of text.
\newblock \emph{arXiv preprint arXiv:1606.05250}.

\bibitem[{Rogoz and Ionescu(2024)}]{rogoz2024unibucllm}
Ana-Cristina Rogoz and Radu~Tudor Ionescu. 2024.
\newblock Unibucllm: Harnessing llms for automated prediction of item difficulty and response time for multiple-choice questions.
\newblock \emph{arXiv preprint arXiv:2404.13343}.

\bibitem[{Team(2024{\natexlab{a}})}]{gemma_2024}
Gemma Team. 2024{\natexlab{a}}.
\newblock \href {https://doi.org/10.34740/KAGGLE/M/3301} {Gemma}.

\bibitem[{Team(2024{\natexlab{b}})}]{qwen2.5}
Qwen Team. 2024{\natexlab{b}}.
\newblock \href {https://qwenlm.github.io/blog/qwen2.5/} {Qwen2.5: A party of foundation models}.

\bibitem[{Team(2025)}]{qwen3technicalreport}
Qwen Team. 2025.
\newblock \href {https://arxiv.org/abs/2505.09388} {Qwen3 technical report}.
\newblock \emph{Preprint}, arXiv:2505.09388.

\bibitem[{Uto et~al.(2023)Uto, Tomikawa, and Suzuki}]{uto2023difficulty}
Masaki Uto, Yuto Tomikawa, and Ayaka Suzuki. 2023.
\newblock Difficulty-controllable neural question generation for reading comprehension using item response theory.
\newblock In \emph{Proceedings of the 18th workshop on innovative use of NLP for building educational applications (BEA 2023)}, pages 119--129.

\bibitem[{Wang et~al.(2022)Wang, Wei, Schuurmans, Le, Chi, Narang, Chowdhery, and Zhou}]{wang2022self}
Xuezhi Wang, Jason Wei, Dale Schuurmans, Quoc Le, Ed~Chi, Sharan Narang, Aakanksha Chowdhery, and Denny Zhou. 2022.
\newblock Self-consistency improves chain of thought reasoning in language models.
\newblock \emph{arXiv preprint arXiv:2203.11171}.

\bibitem[{Wei et~al.(2022)Wei, Wang, Schuurmans, Bosma, Xia, Chi, Le, Zhou et~al.}]{wei2022chain}
Jason Wei, Xuezhi Wang, Dale Schuurmans, Maarten Bosma, Fei Xia, Ed~Chi, Quoc~V Le, Denny Zhou, et~al. 2022.
\newblock Chain-of-thought prompting elicits reasoning in large language models.
\newblock \emph{Advances in neural information processing systems}, 35:24824--24837.

\bibitem[{Xu et~al.(2024)Xu, Li, Sun, and Qian}]{xu2024adaption}
Mayi Xu, Yongqi Li, Ke~Sun, and Tieyun Qian. 2024.
\newblock Adaption-of-thought: Learning question difficulty improves large language models for reasoning.
\newblock In \emph{Proceedings of the 2024 Conference on Empirical Methods in Natural Language Processing}, pages 5468--5495.

\bibitem[{Xu et~al.(2022)Xu, Wang, Yu, Ritchie, Yao, Wu, Zhang, Li, Bradford, Sun et~al.}]{xu2022fantastic}
Ying Xu, Dakuo Wang, Mo~Yu, Daniel Ritchie, Bingsheng Yao, Tongshuang Wu, Zheng Zhang, Toby Jia-Jun Li, Nora Bradford, Branda Sun, et~al. 2022.
\newblock Fantastic questions and where to find them: Fairytaleqa--an authentic dataset for narrative comprehension.
\newblock In \emph{Proceedings of the 60th Annual Meeting of the Association for Computational Linguistics (Volume 1: Long Papers)}, pages 447--460.

\bibitem[{Yang et~al.(2018)Yang, Qi, Zhang, Bengio, Cohen, Salakhutdinov, and Manning}]{yang2018hotpotqa}
Zhilin Yang, Peng Qi, Saizheng Zhang, Yoshua Bengio, William Cohen, Ruslan Salakhutdinov, and Christopher~D Manning. 2018.
\newblock Hotpotqa: A dataset for diverse, explainable multi-hop question answering.
\newblock In \emph{Proceedings of the 2018 Conference on Empirical Methods in Natural Language Processing}, pages 2369--2380.

\bibitem[{Zhou and Tao(2020)}]{zhou2020multi}
Ya~Zhou and Can Tao. 2020.
\newblock Multi-task bert for problem difficulty prediction.
\newblock In \emph{2020 international conference on communications, information system and computer engineering (cisce)}, pages 213--216. IEEE.

\end{thebibliography}

\newpage
\onecolumn

\appendix

\section{Data Annotation Details \label{sec:appendix:annotation_detail}}

\begin{figure*}[t]
    \centering
    \includegraphics[width=\linewidth]{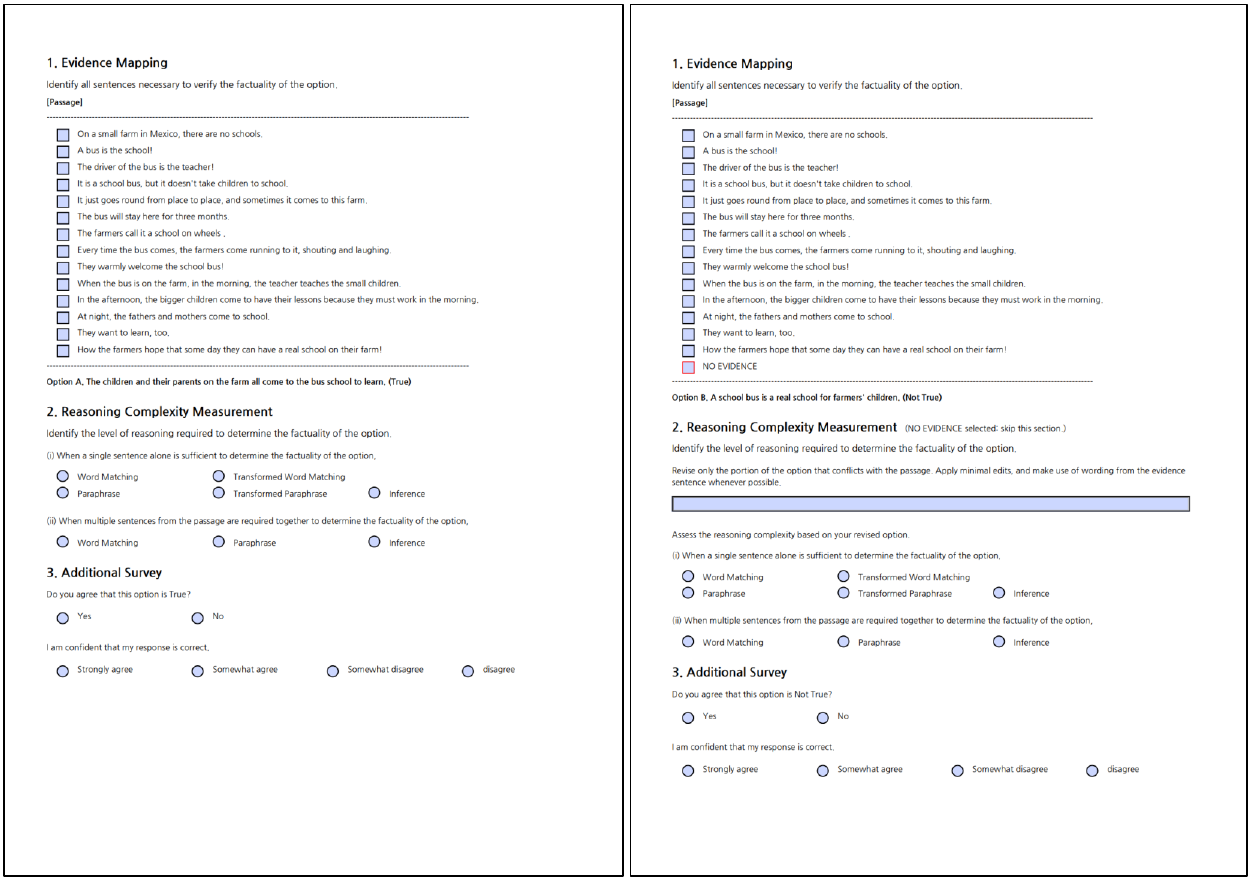}
    \caption{Example annotation sheets for \textit{True} (left) and \textit{Not True} (right) statements. For \textit{Not True} statements, annotators could mark ``No Evidence'' when the passage lacked sufficient information (corresponding to the \textit{Insufficient Evidence} category). Annotators were also asked to create a minimally revised \textit{True} version of each \textit{False} statement to enable assessment of its Transformation Level.}
    \label{fig:annotation_sheet}
\end{figure*}

We recruited three experts via Upwork\footnote{\url{https://www.upwork.com}}, informing them in advance that the task involved data collection for research purposes and that their anonymity would be guaranteed.
They independently labeled 238 MTF items (952 statements) across six batches.
Annotators were compensated per item according to the education level of the source test: \$1.20 for middle school, \$1.50 for high school, and \$2.00 for college-level items, with additional payment for training and revision.

Figure \ref{fig:annotation_sheet} illustrates examples of the annotation sheet used by annotators. 
For each statement, annotators were shown the passage, the statement, and its factuality label.
Annotators first identified the sentence(s) necessary to verify the statement’s factuality (Evidence Scope) and then determined its Transformation Level based on the lexical and structural relationship between the statement and the selected evidence.
Because TL is defined only for \textit{True} statements, annotators revised each \textit{False} statement—excluding those labeled as \textit{Insufficient Evidence}—into its closest \textit{True} version before assigning a TL label.

To further ensure data quality, we added two additional survey questions.
First, annotators were asked whether they agreed with the provided factuality label, and items with disagreement were discarded.
Second, they reported their confidence in the labels they assigned; although no low-confidence items were reported, any such cases would have been excluded.
Finally, items flagged as potentially problematic or ethically inappropriate were also removed.

\begin{figure}[t]
\centering
\includegraphics[width=0.45\columnwidth]{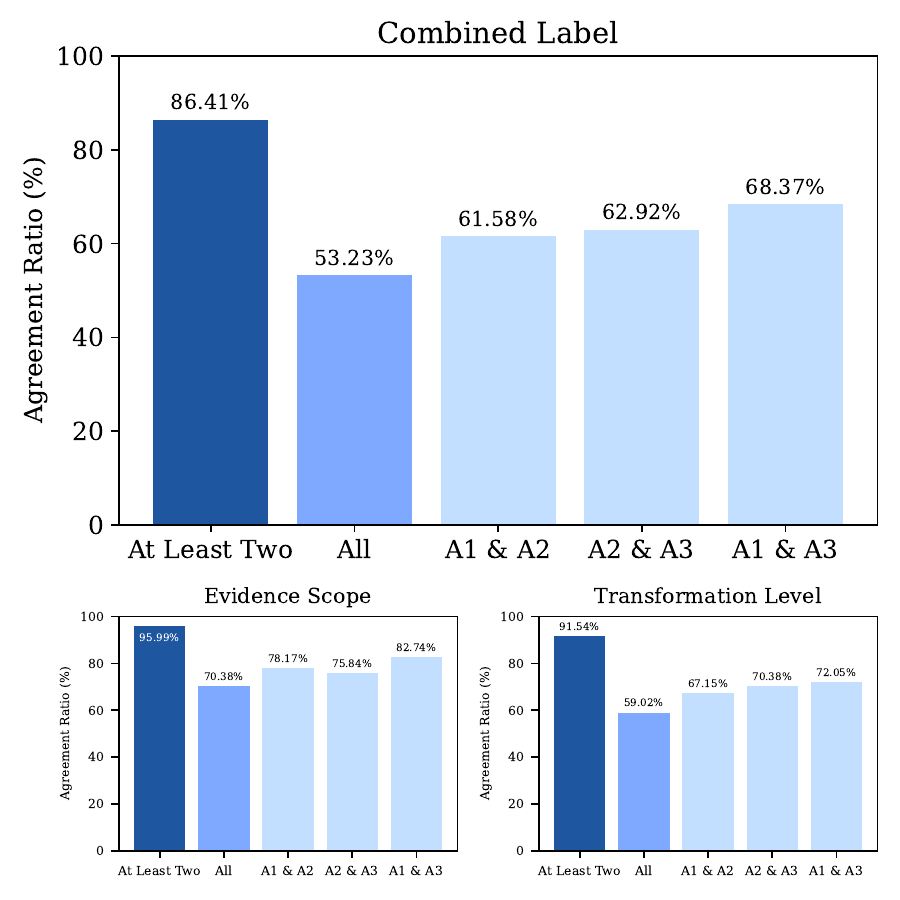}
\caption{\label{fig:agreement} Inter-annotator agreement across labeling dimensions. Agreement ratios are shown for (top) combined labels of two dimensions. ``At Least Two'' indicates majority agreement among annotators, while ``All'' requires unanimous agreement. Pairwise agreements between annotators (A1, A2, A3) are also reported. For TL agreement, multi-evidence items labeled as \textit{word matching} or \textit{paraphrasing} were considered equivalent to \textit{transformed word matching} and \textit{transformed paraphrasing}, respectively.}
\end{figure}

\begin{figure}[h]
\centering
\includegraphics[width=0.65\columnwidth]{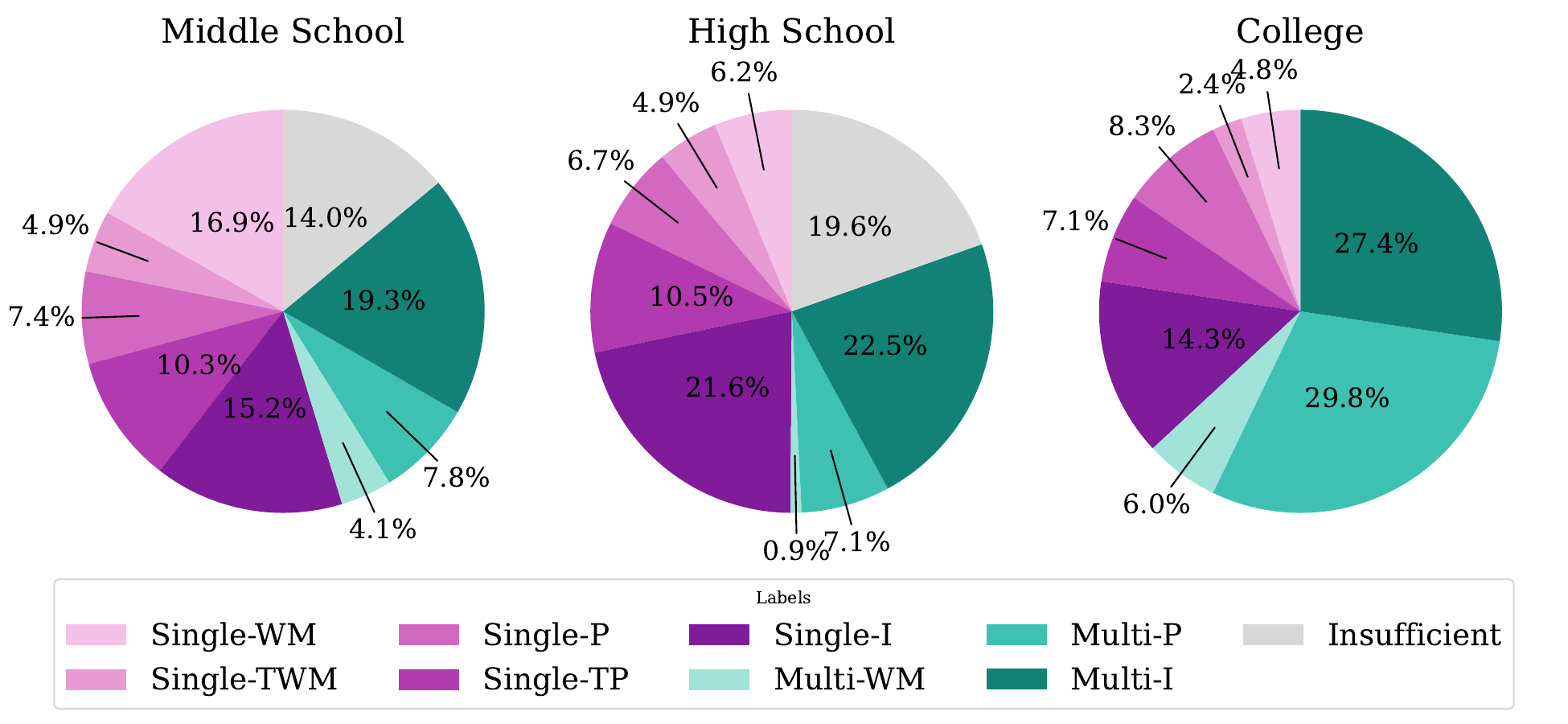}
\caption{\label{fig:label_distrib_test_demo} Distribution of combined difficulty labels in the \dataName\ dataset across educational levels.}
\end{figure}

Figure~\ref{fig:agreement} presents inter-annotator agreement for both annotation dimensions.
Among the 898 items retained after filtering based on annotators’ survey responses, 86.41\% received the same label from at least two annotators, while 53.23\% achieved full agreement among all three.
Although the annotation process was guided by structured label definitions, it required close examination of lexical, syntactic, and inferential relationships between statements and evidence, making it more demanding than typical classification tasks.
In some cases, minor oversights—such as missing subtle paraphrases or nuanced details—led to disagreements.
To mitigate their impact on data quality, we removed items for which all three annotators provided different labels.
For items with partial agreement (i.e., two matching labels and one dissenting), the authors manually reviewed all annotations and resolved discrepancies by cross-referencing annotators’ justifications with the passage content.
After this adjudication process, we obtained 776 annotated TFNG items, which we refer to as \dataName.

\section{\label{sec:appendix:data_analysis} Data Analysis}

\begin{figure}[t]
\centering
\includegraphics[width=0.50\columnwidth]{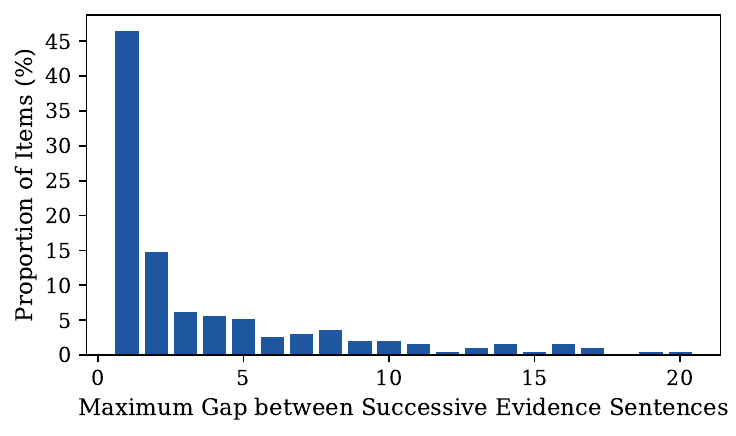}
\caption{\label{fig:evidence_gap} Distribution of the maximum gap between successive evidence sentences for multiple-sentence evidence items. For each item, the gap is measured as the largest distance between any two consecutive evidence sentences in the passage.}
\end{figure}

Figure~\ref{fig:label_distrib_test_demo} presents the distribution of cognitive labels in \dataName\ across different educational levels.
\textit{Word-matching} items with \textit{single-sentence evidence}---representing the lowest cognitive complexity---are more prevalent in middle school exams but decrease substantially at higher levels.
Conversely, \textit{multi-sentence} and \textit{inference}-based items occur more frequently in high school and college assessments.
These patterns suggest that cognitive complexity varies considerably even among items within the same educational level or format, implying that educational level alone is insufficient for fine-grained difficulty analysis.

Figure~\ref{fig:evidence_gap} illustrates the maximum gap between successive evidence sentences, as identified by the raters, for multi-sentence evidence items.
According to the chart, approximately 45\% of these items have evidence sentences that are adjacent in the passage, whereas about 55\% require comprehension across sentences that are farther apart.
This indicates that many items demand integration of information from non-contiguous parts of the passage, rather than relying solely on one or two consecutive sentences.

\section{\label{sec:appendix:experimental_detail} Experimental Details}

In our experiments, we used open-source models from Hugging Face\footnote{\url{https://huggingface.co}} with the following model names:
\begin{itemize}[nosep]
\item Gemma2-9B: \texttt{google/gemma-2-9b-it}
\item Gemma2-27B: \texttt{google/gemma-2-27b-it}
\item Mistral-7B: \texttt{mistralai/Mistral-7B-Instruct-v0.3}
\item Mistral-24B: \texttt{mistralai/Mistral-Small-24B-Instruct}
\item Qwen2.5-7B: \texttt{Qwen/Qwen2.5-7B-Instruct}
\item Qwen2.5-32B: \texttt{Qwen/Qwen2.5-32B-Instruct}
\item Qwen3-32B: \texttt{Qwen/Qwen3-32B}
\end{itemize}
In addition, we used GPT-4o and GPT-4o-mini via the OpenAI API\footnote{\url{https://openai.com}}, with model versions \texttt{gpt-4o-2024-08-06} and \texttt{gpt-4o-mini-2024-07-18}, respectively.

For self-consistency decoding~\citet{wang2022self}, we used the default hyperparameter values of each model that the authors defined, especially for top-$p$, top-$k$, and temperature.
All experiments were conducted using two NVIDIA A100 40GB GPUs.

\section{\label{sec:appendix:micro_performance} Supplementary Micro-F1 Results}

Tables~\ref{tab:main_result_micro}, \ref{tab:subtask_result_accuracy}, and \ref{tab:qwen3_micro} present the Micro-F1 results corresponding to Tables~\ref{tab:main_result}, \ref{tab:subtask_result}, and~\ref{tab:qwen3}, respectively.

\begin{table}[t]
\centering
\small
\begin{tabular}{lccccccccc}
\toprule
\multirow{2}{*}{Method}       & \multirow{2}{*}{\#Demo} & \multicolumn{2}{c}{Gemma2}  & \multicolumn{2}{c}{Mistral} & \multicolumn{2}{c}{Qwen2.5} & \multicolumn{2}{c}{GPT-4o}   \\
                              &                          & 9B        & 27B       & 7B        & 24B       & 7B        & 32B       & mini      & -         \\ \midrule
\multicolumn{10}{c}{\textit{Reading Comprehension}}                                                                                                                                              \\ \midrule
CoT  & 1                        & 85.1     & 89.2     & 63.7     & 79.7     & 80.9     & 88.2     & 87.0     & 89.4     \\ \midrule
\multicolumn{10}{c}{\textit{Evidence Scope Classification} [Human: 87.9]}                                                                                                                                              \\ \midrule
\multirow{3}{*}{\begin{tabular}[c]{@{}l@{}}SP\end{tabular}}  & 0                        & 48.8     & 55.2     & 43.0     & 58.8     & 46.0     & 56.4     & 51.8     & 57.8     \\
                              & 1                        & 48.8     & \underline{60.0}          & 45.8     & \underline{59.8}          & 49.8     & \underline{59.8}          & \underline{55.2}          & 60.8     \\
                              & 6                        & \underline{50.4}          & 53.6     & \underline{49.4}          & 57.0     & \underline{51.8}          & 58.4     & 54.6     & \underline{65.7}          \\ \midrule
\multirow{3}{*}{\begin{tabular}[c]{@{}l@{}}CoT\end{tabular}} & 0                        & 60.0     & 62.7     & 21.3     & 63.7     & 59.2     & 70.5     & 66.3     & 72.5     \\
                              & 1                        & \underline{64.5}          & \underline{70.1}          & 53.6     & 66.3     & 58.4     & \textbf{\underline{73.1}} & \underline{68.9} & \textbf{\underline{75.5}} \\
                              & 6                        & 62.9     & 69.1     & \underline{55.6}          & \underline{68.5}          & \underline{60.6}          & 70.7     & 66.7     & 69.5     \\ \midrule
\multirow{3}{*}{\begin{tabular}[c]{@{}l@{}}CoT (SC)\end{tabular}}   & 0                        & 61.7     & 67.9     & 19.5     & 68.3     & 60.4     & \textbf{\underline{73.1}} & 67.3               & 72.9               \\
                              & 1                        & \textbf{\underline{67.7}} & \textbf{\underline{72.3}} & \textbf{\underline{57.8}} & \textbf{\underline{71.1}} & \textbf{\underline{61.9}} & 72.9     & \textbf{\underline{72.3}}               & \underline{74.1}               \\
                              & 6                        & 63.9     & 70.7     & 56.6     & 68.9     & 60.8     & 71.5     & 69.3               & 72.9               \\ \midrule

\multicolumn{10}{c}{\textit{(3-level) Transformation Level Classification} [Human: 85.9]}                                                                                                                              \\ \midrule
\multirow{3}{*}{\begin{tabular}[c]{@{}l@{}}SP\end{tabular}}  & 0                        & 57.9     & 55.9     & 46.8     & 73.7     & \underline{62.4}          & 67.5     & 67.0     & \underline{70.7}          \\
                              & 1                        & 55.2     & 58.3     & \underline{58.4}          & \underline{75.4}          & 58.8     & \underline{69.0}          & \underline{67.8}          & 69.7     \\
                              & 8                        & \underline{58.3}          & \textbf{\underline{64.3}} & 55.7     & 73.5     & 54.7     & 60.0     & 59.4     & 64.9     \\ \midrule
\multirow{3}{*}{\begin{tabular}[c]{@{}l@{}}CoT\end{tabular}} & 0                        & 53.1     & 58.8     & 59.0     & \underline{76.9}          & 68.6     & 72.4     & 67.8     & 76.6     \\
                              & 1                        & 50.9     & 56.8     & 48.6     & 75.2     & \underline{76.2} & \underline{74.7}          & 71.9     & \textbf{\underline{78.1}} \\
                              & 8                        & \underline{57.1}          & \underline{61.3}          & \underline{62.8}          & 75.4     & 66.0     & 73.2     & \textbf{\underline{73.6}} & 75.4     \\ \midrule
\multirow{3}{*}{\begin{tabular}[c]{@{}l@{}}CoT (SC)\end{tabular}}   & 0                        & 58.6     & 54.6     & \textbf{\underline{68.5}} & \textbf{\underline{83.2}} & \textbf{\underline{76.5}} & \textbf{\underline{78.3}}  & 69.6               & 74.1               \\
                              & 1                        & 54.6     & 58.0     & 52.2     & 67.9     & 72.3     & 76.2     & \underline{71.5}               & \underline{74.3}               \\
                              & 8                        & \textbf{\underline{64.2}} & \underline{62.7}          & 64.1     & 76.9     & 66.8     & 73.3     & 71.4               & 71.8               \\ \midrule

\multicolumn{10}{c}{\textit{(5-level) Transformation Level Classification} [Human: 83.5]}                                                                                                                              \\ \midrule
\multirow{3}{*}{\begin{tabular}[c]{@{}l@{}}SP\end{tabular}}  & 0                        & 36.8     & 38.8     & 28.4     & 52.0     & 43.6     & 51.2     & 51.2     & \underline{54.8}          \\
                              & 1                        & \underline{42.0}          & 42.4     & 34.0     & \underline{57.2}          & \underline{47.6}          & \underline{52.0}          & \underline{50.8}          & 52.8     \\
                              & 8                        & 35.6     & \textbf{\underline{49.6}} & \underline{40.0}          & 56.4     & 42.8     & 43.2     & 48.8     & 52.4     \\ \midrule
\multirow{3}{*}{\begin{tabular}[c]{@{}l@{}}CoT\end{tabular}} & 0                        & 34.0     & 43.2     & 38.8     & \underline{59.2}     & 42.8     & \underline{59.6}     & 52.8     & 66.0     \\
                              & 1                        & 37.6     & 38.4     & 32.0     & 54.0     & \underline{53.6} & 58.4          & 56.0     & \textbf{\underline{67.2}} \\
                              & 8                        & \underline{42.8}          & \underline{47.6}          & \underline{48.0}          & 56.4          & 48.0     & 55.6     & \textbf{\underline{58.8}} & 59.2     \\ \midrule
\multirow{3}{*}{\begin{tabular}[c]{@{}l@{}}CoT (SC)\end{tabular}}   & 0                        & 37.6     & 44.0     & 44.8     & \textbf{\underline{67.2}} & 46.0     & \textbf{\underline{66.0}} & 54.8               & 60.4               \\
                              & 1                        & 43.2     & 45.2     & 39.2     & 53.6     & \textbf{\underline{56.8}} & 63.2     & 55.6               & \underline{63.6}               \\
                              & 8                        & \textbf{\underline{48.0}} & \underline{48.4}          & \textbf{\underline{50.0}} & 58.8     & 45.2     & 59.2     & \underline{56.4}               & 58.8               \\ \bottomrule
\end{tabular}
\caption{\label{tab:main_result_micro} 
Micro-F1 scores on the RC task and the ES and TL classification tasks.
Greedy decoding is the default inference method, and SC indicates that self-consistency decoding is used.
\textbf{Bolded} values denote each model’s best score per task; \underline{underlined} values indicate the best score across demonstration settings.}
\end{table}

\begin{table}[t]
\centering
\small
\begin{tabular}{lrrrrrrrr}
\toprule
\multirow{2}{*}{Subtask}            & \multicolumn{2}{c}{Gemma2}                       & \multicolumn{2}{c}{Mistral}                      & \multicolumn{2}{c}{Qwen2.5}                     & \multicolumn{2}{c}{GPT-4o}                       \\
                                 & \multicolumn{1}{c}{9B} & \multicolumn{1}{c}{27B} & \multicolumn{1}{c}{7B} & \multicolumn{1}{c}{24B} & \multicolumn{1}{c}{7B} & \multicolumn{1}{c}{32B} & \multicolumn{1}{c}{mini} & \multicolumn{1}{c}{-} \\ \midrule
1.1. Falsifiability Judgment     & 82.8                   & 82.1                   & 71.4                  & 81.7                   & 77.5                  & \textbf{90.1}          & 81.3                     & 87.8                 \\
1.2. Evidence Sentence Counting        & 68.7                    & \textbf{74.3}          & 59.2                  & 69.4                   & 68.5                  & 73.8                   & 67.0                    & 71.9                 \\ \midrule
2.1. Inference Detection         & 73.4                   & 75.8                   & 65.5                  & \textbf{82.4}           & 68.7                   & 81.7                   & 81.4                    & 80.0                 \\
2.2. Paraphrasing Detection      & 81.9                   & 84.9                   & 67.3                  & 85.9                   & 72.4                  & 86.4                   & 88.4                    & \textbf{88.9}        \\
2.3. Phrase Reordering Detection & 59.5                   & \textbf{68.0}          & 58.8                  & 61.4                   & 51.0                  & 54.3                   & 66.7                    & 65.4    \\ \bottomrule            
\end{tabular}
\caption{\label{tab:subtask_result_accuracy} Micro-F1 scores on subtasks that measure fine-grained abilities required for the main classification tasks. \textbf{Bolded} values denote the best model for each subtask.}
\end{table}

\begin{table}[t]
\centering
\small
\begin{tabular}{lccc}
\toprule
Model          & ES             & \begin{tabular}[c]{@{}c@{}}TL\\(\textit{5-level})\end{tabular}   & \begin{tabular}[c]{@{}c@{}}TL\\(\textit{3-level})\end{tabular} \\ \midrule
Qwen2.5-32B    & 70.5           & 59.6       & 72.4 \\
Qwen3-32B$_{non-thinking}$      & 67.9        & 63.2       & 78.9 \\
Qwen3-32B$_{thinking}$   & 65.5        & 60.4      & 71.8  \\ \bottomrule
\end{tabular}
\caption{\label{tab:qwen3_micro} Comparison of Micro-F1 scores for LLMs with and without deep-thinking capabilities.}
\end{table}

\end{document}